\documentclass[10pt,twocolumn,letterpaper]{article}

\usepackage{cvpr}              
\usepackage{amsmath}
\usepackage{amssymb}
\usepackage{slashbox}
\usepackage{mathtools}
\usepackage{amsthm}
\usepackage{amsmath,amsfonts,bm}
\usepackage{times}
\usepackage{epsfig}
\usepackage{graphicx}
\usepackage{amsmath}
\usepackage{amssymb}
\usepackage{float}
\usepackage{tabularx}
\usepackage{rotating}

\newenvironment{squishlist}
{   \begin{list}{$\bullet$}
    { 
    \setlength{\itemsep}{0pt}
    \setlength{\parsep}{.5em}
    \setlength{\topsep}{0pt}
    \setlength{\partopsep}{0pt}
    \setlength{\leftmargin}{1.5em} 
    \setlength{\labelwidth}{1.5em}
    \setlength{\labelsep}{0.8em} } }
      {\end{list}}
\theoremstyle{plain}

\theoremstyle{definition}

\theoremstyle{remark}

\usepackage[textsize=tiny]{todonotes}
\usepackage{makecell}
\newcommand{\HHLINE}{\Xhline{1.5\arrayrulewidth}}

\def\eg{\emph{e.g.,}}           

\def\eqref#1{equation~\ref{#1}}

\def\1{\bm{1}}

\def\vx{{\bm{x}}}

\def\mA{{\bm{A}}}

\DeclareMathAlphabet{\mathsfit}{\encodingdefault}{\sfdefault}{m}{sl}
\SetMathAlphabet{\mathsfit}{bold}{\encodingdefault}{\sfdefault}{bx}{n}

\definecolor{cvprblue}{rgb}{0.21,0.49,0.74}
\usepackage[pagebackref,breaklinks,colorlinks,allcolors=cvprblue]{hyperref}

\usepackage[capitalize,noabbrev]{cleveref}

\usepackage[table]{xcolor}

\newcommand\Tstrut{\rule{0pt}{2.6ex}}
\newcommand\Bstrut{\rule[-0.9ex]{0pt}{0pt}}

\def\paperID{} 
\def\confName{CVPR}
\def\confYear{2026}
\newcommand{\equalcontrib}{\textsuperscript{*}}
\newcommand{\corresp}{\textsuperscript{\dag}}
\title{Residual Connections Harm Generative Representation Learning}

\author{
    Xiao Zhang\equalcontrib$^{1,3}$ \quad 
    Ruoxi Jiang\equalcontrib\corresp$^{1,2,4}$ \quad 
    William Gao$^{1}$ \quad 
    Rebecca Willet$^{1}$ \quad 
    Michael Maire$^{1}$
    \\[1.5ex]
    \small $^1$University of Chicago \quad $^2$Fudan University \quad $^3$ Tencent
    \quad $^4$ Shanghai Academy of AI for Science
}

\begin{document}
\maketitle
\newcommand{\btheta}{\mathbf{\theta}}

\renewcommand{\thefootnote}{\fnsymbol{footnote}}
\footnotetext[1]{Both authors contributed equally to this work. This work was completed while Xiao Zhang was a graduate student at Unversity of Chicago.} 
\footnotetext[2]{Corresponding author: roxie\_jiang@fudan.edu.cn} 

\makeatletter
\def\@makefnmark{}     
\makeatother

\renewcommand{\thefootnote}{\arabic{footnote}} 
\footnotetext{Link to code: \url{https://github.com/xiao7199/decayed_Identity_shortcuts}}



\begin{abstract}
We show that introducing a weighting factor to reduce the influence of identity shortcuts in residual networks significantly enhances semantic feature learning in generative representation learning frameworks, such as masked autoencoders (MAEs) and diffusion models.  Our modification notably improves feature quality, raising ImageNet-1K K-Nearest Neighbor accuracy from 27.4\% to 63.9\% and linear probing accuracy from 67.8\% to 72.7\% for MAEs with a ViT-B/16 backbone, while also enhancing generation quality in diffusion models. This significant gap suggests that, while residual connection structure serves an essential role in facilitating gradient propagation, it may have a harmful side effect of reducing capacity for abstract learning by virtue of injecting an echo of shallower representations into deeper layers.  We ameliorate this downside via a fixed formula for monotonically decreasing the contribution of identity connections as layer depth increases.  Our design promotes the gradual development of feature abstractions, without impacting network trainability.  Analyzing the representations learned by our modified residual networks, we find correlation between low effective feature rank and downstream task performance. 
\end{abstract}

\vspace{-5pt}
\section{Introduction}
\label{sec:intro}

\begin{figure*}[t]
   \centering
   \includegraphics[width=.90\linewidth]{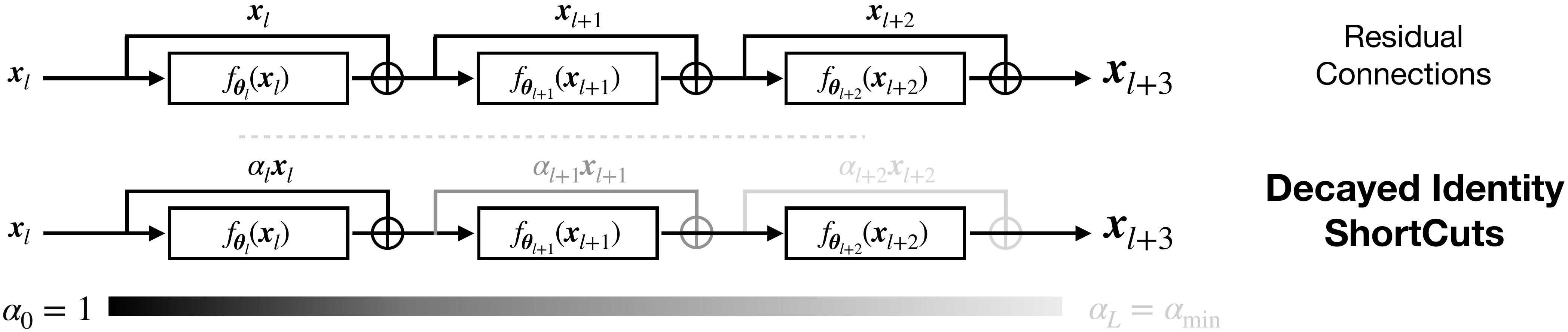}
   \caption{%
      Our \emph{decayed identity shortcuts} introduce a depth-dependent scaling factor to shortcuts in a residual
      network, thereby modulating the contribution of preceding layers and fostering greater abstraction in deeper
      layers.  A simple schema for controlling decay factor $\alpha$ suffices to improve feature learning in both
      MAEs and diffusion models, as well as diffusion model generation quality.
   }%
   \label{fig:teaser}
\vspace{-5pt}
\end{figure*}

Residual networks (ResNets)~\citep{he2016deep} define a connection structure that has achieved near-universal adoption
into modern architectures for deep learning.  At the time of their development, supervised learning
(\eg~ImageNet~\citep{deng2009imagenet} classification) was the driving force behind the evolution of convolutional
neural network (CNN) architectures.  Residual networks solved a key issue: CNNs constructed of more than approximately
20 convolutional layers in sequence became difficult to train, leading to shallower networks outperforming deeper
ones, unless additional techniques, such as auxiliary outputs~\citep{szegedy2015going} or batch
normalization~\citep{ioffe2015batch}, were employed.  Both ResNets, and their predecessor, highway
networks~\citep{srivastava2015highway} provide elegant solutions to this trainability problem by endowing the network
architecture with alternative shortcut pathways along which to propagate gradients.  Highway networks present a more
general formulation that modulates these shortcut connections with learned gating functions.  However, given their
sufficient empirical effectiveness, the simplicity of ResNet's identity shortcuts makes them a
preferred technique.

While solving the gradient propagation issue, residual connections impose a specific functional form on the network;
between residual connections, each layer (or block of layers) learns to produce an update slated to be added to its
own input.  This incremental functional form may influence the computational procedures learned by the
network~\citep{greff2016highway}.  Alternatives to residual and highway networks exist that do not share this
functional form, but implement other kinds of skip-connection scaffolding in order to assist gradient
propagation~\citep{larsson2016fractalnet,huang2017densely,zhu2018sparsely}.  Thus, shortcut pathways, rather than a
specific form of skip connection, are the essential ingredient to enable the training of very deep networks.
Nevertheless, nearly all modern large-scale models, including those based on the transformer
architecture~\citep{vaswani2017attention} incorporate the standard residual connection.

This design choice holds, even as deep learning has shifted into an era driven by self-supervised training.  The
shift to self-supervision brings to the forefront new learning paradigms, including those based on contrastive~\citep{
wu2018unsupervised,he2020momentum,chen2020simple,caron2021emerging,grill2020bootstrap}, generative~\citep{
goodfellow2014generative,DBLP:journals/pami/KarrasLA21,DBLP:conf/nips/HoJA20,DBLP:conf/iclr/SongME21,
song2023consistency,rombach2022high}, and autoencoding~\citep{kingma2013auto,he2022masked,li2023mage} objectives.
Many systems in the generative and autoencoding paradigms rely on ``encoder-decoder'' architectures, often styled
after the original U-Net~\citep{ronneberger2015u}, which contains additional long-range shortcuts between corresponding
layers in mirrored symmetry about a central bottleneck.  With representation learning as a goal, one typically desires
that the middle bottleneck layer produce a feature embedding reflecting abstract semantic properties.  The interaction
of skip-connection scaffolding for gradient propagation with encoder-decoder architectures, self-supervised training
objectives, and bottleneck representations has not been carefully reconsidered.  This is a worrisome oversight,
especially since, even in the supervised setting with standard classification architectures, prior work suggests that
unweighted identity shortcuts may be a suboptimal design decision~\citep{savarese2017residual,fischer2023optimal}.
Intuitively, identity shortcuts may not be entirely appropriate for capturing high-level, semantic
features as they directly inject low-level, high-frequency details of inputs into outputs, potentially
compromising feature abstraction.  We explore this issue within generative learning frameworks, including
masked autoencoders (MAEs)~\citep{he2022masked} and diffusion models~\citep{DBLP:conf/nips/HoJA20}, leading paradigms
for self-supervised image representation learning and generation.  Our experiments demonstrate that identity shortcuts
significantly harm semantic feature learning in comparison to an alternative we propose: gradually decay the weight of
the identity shortcut over the depth of the network, thereby reducing information flow through it
(Figure~\ref{fig:teaser}).  With increasing layer depth, our approach facilitates a smooth transition from a residual
to a feed-forward architecture, while maintaining sufficient connectivity to train the network effectively.  Unlike
prior work on learned gating~\citep{srivastava2015highway} or reweighting~\citep{savarese2017residual} mechanisms for
residual connections, our method is a forced decay scheme governed by a single hyperparameter.

\begin{figure*}[t]
   \centering
   \includegraphics[width=.99\linewidth]{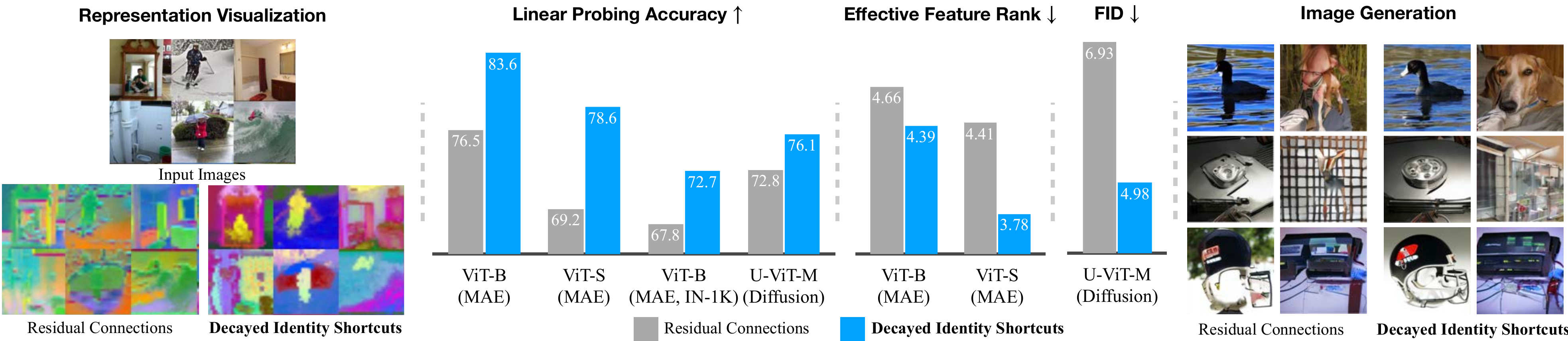}
   \caption{%
      We design \emph{decayed identity shortcuts}~(Figure~\ref{fig:teaser}), a variant of residual connections, to
      facilitate self-supervised representation learning in generative model.  Compared to standard residual connections, our approach
      yields superior abstract semantic features (\emph{left}, visualized using \citet{zhang2024deciphering}'s
      approach), whose leading components pop out object instances and classes.
      Quantitative evaluation shows our
      architecture encourages lower feature rank and learns better feature representation for
      both MAE and diffusion models (\emph{middle}), along with enhanced generation quality for diffusion models (\emph{right}).
      These improvements require no additional learnable parameters.%
   }%
   \label{fig:teaser_intro}
\end{figure*}

A parallel motivation for our design stems from \citet{huh2021low}, who show that features from residual
blocks have higher rank than those produced by comparative feed-forward blocks.  The smooth transition between
residual and feed-forward behavior induced by our decay scheme regularizes deeper features toward exhibiting
low-rank characteristics.  Section~\ref{sec:discussion} experimentally explores the correlation between our decayed
identity shortcuts and low-rank feature representations.  Figure~\ref{fig:teaser_intro} previews the corresponding
improvements to feature learning.  Our contributions are:

\begin{squishlist}
\item We introduce decayed identity shortcuts, a simple architectural mechanism which enhances feature
abstraction in masked autoencoders and diffusion models. 
\item Our design within an MAE yields a substantial performance boost on
ImageNet-1K~\citep{deng2009imagenet}: achieving a linear probing accuracy of 72.7\% (up from a baseline of 67.8\%) and a K-Nearest Neighbor accuracy of 63.9\% (an improvement from the baseline of 27.4\%).
\item In diffusion models, our design improves both feature learning and generation quality.
\item Smaller models with decayed identity shortcuts outperform larger ones using standard residual connections.
\end{squishlist}

\section{Related Work}

\textbf{Self-supervised representation learning.}
Recent advancements~\citep{achiam2023gpt,kirillov2023segment,rombach2022high,team2023gemini,shi2020improving,
ramesh2021zero} in deep learning follow a common scaling law, in which a model's performance consistently improves
with its capacity and the size of the training data.  This effect can be observed in large language models (LLMs),
which are trained on vast amounts of internet text, enabling them to perform some tasks at a human
level~\citep{laskar2023systematic} and exhibit remarkable zero-shot capabilities~\citep{kojima2022large}.  These models
are trained using next-token-prediction, allowing them to be trained without labeled data.  In contrast, the progress
of this scaling law in computer vision has largely depended on annotated data.  For instance, the Segment Anything
~\citep{kirillov2023segment} leverages 1 billion human-annotated masks, and state-of-the-art image
generators~\citep{ramesh2021zero} require training on huge datasets of text-image pairs~\citep{schuhmann2022laion}.
However, the vast volume of unlabeled visual data and desire for continued scaling motivates a transition to
self-supervised learning.

At present, two families of approaches to self-supervised visual representation learning appear particularly
promising.

\noindent\textbf{Contrastive representation learning}~\citep{wu2018unsupervised,he2020momentum,chen2020simple,caron2021emerging,
grill2020bootstrap}
achieves state-of-the-art performance in most downstream classification tasks by training discriminative models to maximize mutual information between differently augmented views of images. However, these approaches rely on extensive and intricate data augmentation pipelines, necessitating domain expertise for adaptation to new domains.

\noindent\textbf{Generative representation learning}, via masked image modeling~\citep{bao2021beit,he2022masked,chen2024context}, which trains to
reconstruct occluded pixels, or via diffusion denoising~\citep{song2020score,DBLP:conf/nips/HoJA20,
DBLP:conf/iclr/SongME21}, which trains to reverse a process that mixes images with Gaussian noise, relying less on forming discriminative augmentations, learns to extract representations inherently along the generative process.
Some hybrid approaches~\citep{zhou2021ibot,huang2023contrastive,li2023mage} combine both families.  Despite advancements, neither
has demonstrated the same scalability~\citep{singh2023effectiveness} as seen in LLMs.  This challenge is additional
motivation for reconsidering the foundations of self-supervised network architectures.

\noindent
\textbf{Residual and skip-connection architectures.}
Highway networks~\citep{greff2016highway} first propose an additive skip connection structure to provide a scaffolding
for gradient propagation when training very deep (\eg~100 layer) networks.  Motivated by the gating mechanisms within
LSTMs~\citep{hochreiter1997long}, this solution uses learned gating functions to weight each combination of identity
and layer output branches.  Residual networks~\citep{he2016deep} are a simplification that removes these learned
coefficients.  DenseNet~\citep{huang2017densely} and FractalNet~\citep{larsson2016fractalnet} demonstrate that access
to gradient paths of multiple lengths is the core requirement of training scaffolding, by introducing skip-connection
structures with other functional forms.  DenseNet utilizes feature concatenation instead of addition, while FractalNet
imposes a recursive tree-like architecture combining subnetworks of multiple depths.

\citet{zhu2018sparsely} explore variants of ResNets and DenseNets with fewer points of combination between different
internal paths, demonstrating that a sparser scaffolding structure may be more robust as network depth increases to
thousands of layers.  \citet{savarese2017residual} add a scalar gating functional to the layer output in residual
networks, yielding a hybrid design between residual and highway networks; learning this scalar gating provides a
consistent benefit to classification accuracy.  \citet{fischer2023optimal} develop a weighting scheme for residual
connections based upon a sensitivity analysis of signal propagation within a ResNet.  To date, none of these potential
improvements has seen broad adoption.

\noindent
\textbf{Low rank bias in neural networks.}
Over-parameterized neural networks exhibit surprising generalization capabilities, a finding seemingly in contradiction
with classical machine learning theory \citep{neyshabur2018towards}.  This phenomenon implies the existence of some form
of implicit regularization that prevents the model from overfitting.  From the perspective of model 
parameterizations, \citet{arora2019implicit} suggest that linear models with more layers tend to converge to minimal
norm solutions.  In the context of CNNs, \citet{huh2021low} demonstrate that stacking more feed-forward layers compels
the model to seek lower rank solutions,  and \citet{jing2020implicit} reinforce this finding by adding more layers
to an autoencoder's bottleneck, thereby creating a representation bottleneck.  In vision transformers,
\citet{geshkovski2024emergence} examine the connection between attention blocks and mean-shift clustering
\citep{cheng1995mean}, showing that repeated attention operations result in low-rank outputs.  Moreover,
\citet{dong2021attention} reveal that eliminating the shortcut connection from residual attention blocks causes
features to degenerate to rank 1 structures doubly exponentially.  From a different perspective, recent
work~\citep{radhakrishnan2022feature,beaglehole2023mechanism, radhakrishnan2024linear} shows training algorithms
implicitly induce low-rank behavior in neural networks.  \citet{radhakrishnan2024linear} study the dimensionality
reduction behavior of a recursive feature machine~\citep{radhakrishnan2022feature} and effectively verify performance
on low-rank matrix recovery.

\section{Method}
\label{sec:method}
In this section, we present the methodology for promoting a low-rank inductive bias using our proposed decay schema and discuss additional implementation strategies designed to stabilize the training process.

\subsection{Decayed Identity Shortcuts}
\label{sec:reduc_resid}
\textbf{Feed-forward layers.} Consider a neural network of $L$ layers. 
For each layer $l$ parameterized with $\btheta_l$, the operation of a feed-forward neural network can be described as:
\begin{equation}
    \vx_{l+1} = f_{\btheta_l}(\vx_l),
    \label{eqn:feedforward}
\end{equation}
where $\vx_l \in \mathbb{R}^d$ represents the output from the preceding layer, and $f_{\btheta_l}$ denotes the network block applied at the current layer. Although it is widely known that pure feed-forward architectures are susceptible to vanishing gradients when building deeper models, \citet{huh2021low} demonstrates that feed-forward modules offer implicit structural regularization, enabling deep models to generate abstract representations at bottlenecks.

\noindent
\textbf{Residual connections.}
To address the optimization problem of vanishing gradients in deeper neural networks,
ResNets~\citep{he2016deep} construct each layer as a residual function, resulting in a modification to Eq.~\ref{eqn:feedforward}:
\begin{equation}
    \vx_{l+1} = \vx_l + f_{\btheta_l}(\vx_l).
\end{equation}
This design builds shortcuts from input to output, allowing gradient magnitude to be preserved regardless of the depth of the model. However, a consequence of this design is that the output stays close to the input in practice \citep{greff2016highway}, defeating the need to construct complex transformations over depth.  The same phenomenon is also observed in highway networks \citep{srivastava2015highway}, which adopt learnable gates $H_{\phi}(\vx) \in [0,1]^{d}$ in both the residual and skip branches:
$\vx_{l+1} = H_{\phi}(\vx_l)\cdot \vx_l + (1 - H_{\phi}(\vx_l))\cdot f_{\btheta_l}(\vx_l)$. Although this flexible design allows the model to build the abstraction level over depths, similar to feedforward networks, \citet{srivastava2015training} finds $H_{\phi} \approx 1$ for most units, suggesting the model prefers copying the input. 

\noindent
\textbf{Decayed identity shortcuts for unsupervised representation learning.}
Setting aside the optimization benefits brought by residual connections,
we rethink the role of the residual connections from the viewpoint of representation learning. Abstraction can be viewed as invariance to local changes of input and is crucial to the disentanglement of the feature space \citep{bengio2013representation}. Prior work suggests that a shortcut path of residual connections tends to preserve high-frequency fine-grained input information \citep{greff2016highway}, resulting in decreased feature abstraction. We hypothesize that this lack of abstraction harms the capability of the model to learn meaningful low-level features and that ensuring an abstract structure in the deeper layers of the neural network will help improve representation learning, especially for unsupervised tasks that often use indirect proxy objectives, such as pixel-wise reconstruction loss. Motivated by this hypothesis, we propose to downweight the contribution from the shortcut path:
\begin{equation}
    \vx_{l+1} = \alpha_l \vx_l + f_{\btheta_l}(\vx_l),
\end{equation}
where $\alpha_l \in [0,1]$ is a rescaling factor to the residual path, controlling the information flow through the skip connection. Fully expanding this relation for a network with $L$ layers indexed from 1 to $L$, we have that:
\begin{equation}
    \vx_{L+1} = \left(\prod_{l=1}^{L} \alpha_l\right)\vx_0 + \sum_{l=1}^{L-1}\left(\prod_{i=l+1}^{L}\alpha_i\right)f_{\btheta_l}(\vx_l) + f_{\btheta_{L}}(\vx_{L}).
\end{equation}
We see that the contribution of the input $\vx_0$ is scaled by each $\alpha_l \leq 1$ while each subsequent network block output $f_{\btheta_l}(\vx_l)$ omits scaling factors up to $\alpha_l$. Hence, the contribution of early features of the network is especially down-weighted, preventing the network from passing fine-grained detailed information to the bottleneck $X_{L+1}$.
 During our experiments, we find the effective decay factor of the final layer, $\alpha_L^{\rm eff}  = \prod_{l = 1}^L \alpha_l$,
plays a critical role in deciding the optimal decay rate when varying the network depth $L$.

\noindent
\textbf{Decay schema.} Instead of specifying $\alpha_l$ as a constant across all layers, we choose $\alpha_l$ to be a function parameterized by the layer index $l$, where the contribution from the shortcut path is monotonically decreasing when $l$ increases:
\begin{equation}
    \alpha_l = 1-\delta_{\alpha}l,
\end{equation}
where $\delta_{\alpha}:= \frac{(1-\alpha_{\rm min})}{L}$, $\alpha_{L} \equiv \alpha_{\rm min}$ is a minimum scaling factor applied at the final layer $L$. Our formulation brings two primary benefits. First, $\alpha_l$, as a linear interpolation between 0 and 1, acts as a smooth transition between residual connections and feedforward layers,
bringing us the optimization benefits from residual connections, while simultaneously encouraging the deeper layers to learn more abstract representations.
Second, similar to the naive formulation, our method only introduces one extra hyperparameter $\alpha_{\rm min}$, which is not data-dependent and does not need to be learned.

\subsection{Implementation Strategy}
\label{sec:method_impl}

\noindent
\textbf{Skip connections for autoencoders.}
Since our method progressively decays the residual connections over network depth, it encourages the most abstract features to be learned by later layers. 
However, learning an abstract bottleneck is detrimental to the training objectives that aim for pixel-wise reconstruction, 
as they necessitate the preservation of detailed information. 
To address this, we incorporate standard skip connections between the encoder and decoder, enabling the encoder to directly pass information 
from shallow layers to the decoder while learning increasingly abstract representations in the deeper encoder layers. 

\noindent
\textbf{Stabilizing training with residual zero initialization.}
The model exhibits rapid feature norm growth at the beginning of training for $\alpha_{\rm min} \leq 0.7$. 
We suspect that the model learns to amplify the output feature norm of $f_{\btheta_l}(\vx)$ to counteract the significant decay applied to the residual connection. 
This growth leads to training instability and negatively impacts training convergence. To address this issue, we follow the implementation of previous works~\citep{DBLP:conf/nips/HoJA20} 
and initialize the weights of the final output layer in each $f_{\btheta_l}$ to zero instead of using the original Xavier uniform initialization \citep{glorot2010understanding}.
This approach enhances training stability by controlling the growth of feature norm, especially with smaller $\alpha_{\rm min}$.

\section{Experiments on Masked Autoencoders}
\label{sec:experiments_mae}
For masked autoencoders (MAEs) \citep{he2022masked}, we replace the residual connections in the encoder's MLP and attention blocks with decayed identity shortcuts. 
The MAE operates by accepting images with a random subset of pixels masked out and learning to recover the discarded pixels. Since the original MAE has twice the number of encoder layers as decoder layers, we build encoder-decoder skip connections by injecting output 
from every other encoder layer into the corresponding decoder layer.  To match spatial dimensions, injected encoder features are combined 
with learnable masked tokens before channel-wise concatenation. The implementation details for the training and evaluation are shown in 
Section~\ref{sec:appendix_experiments_details}. \citet{he2022masked} show the desired representations appear at the end of encoder; we therefore
apply our decaying schema only to the encoder. 

\subsection{Representation Learning on ImageNet-1k}
We follow the default hyperparameters from MAE~\cite{he2022masked} to pretrain ImageNet-1K train split \citep{deng2009imagenet} and use the standard protocol to evaluate the learned representation with end-to-end finetuning (FT), linear probing (LP) and K-Nearest Neighbour (KNN, K = 20), for image classification task. Please see the appendix for detailed experimental setups. 

\begin{table}[t]
    \centering
    \small
        \begin{tabular}{lccc}
         \HHLINE 
         \textbf{Method} & \textbf{FT} & \textbf{LP} & \textbf{KNN} \\
         \hline
         \multicolumn{3}{l}{\textbf{\textit{Contrastive representation learning}}} &\\
         MoCo-v3\citep{chen2021empirical} & 83.2 & 76.7 &  66.6 \\
         DINO\citep{caron2021emerging}  & 83.3 & \textbf{78.2} & \textbf{76.1} \\
         Con MIM \citep{yi2022masked} &  \textbf{83.7} & 39.3 &  - \\
         \hline
         \multicolumn{3}{l}{\textbf{\textit{Generative representation learning}}} & \\
         Data2Vec\citep{baevski2022data2vec} &  84.2 & 68.0 & 33.2 \\
         I-JEPA\citep{assran2023self} & - &\textbf{72.9}  & - \\
         CAE\citep{chen2024context}   & 83.8 & 70.4 & 51.4 \\
         ADDP(VIT-L) \citep{tian2023addp}  &  \textbf{85.9} & 23.8 & -\\
         Latent MIM\citep{wei2024towards}  & 83.0 & 72.0 & 50.1\\
         \rowcolor{gray!15}
         MAE\citep{he2022masked} & 83.6 & 67.8 & 27.4 \\
         \rowcolor{gray!15}
         MAE ($\alpha_{\rm min}$ = 0.6)  & 82.9 &\bf 72.7 & \textbf{63.9}\\
         \HHLINE
    \end{tabular}
    \caption{\textbf{Benchmark of representations on the ImageNet-1K with ViT-B.} 
    Evaluate learned features using standard evaluation protocols: linear probing (LP), fine-tuning (FT) and K-Nearest Neighbor (KNN). 
    With only a simple architectural modification to MAE \citep{he2022masked} and trained purely with pixel-wise reconstruction loss, we achieve 72.7\% LP accuracy and 63.9\% KNN accuracy, significantly narrowing down the gap between generative and contrastive representation learning frameworks}
    \label{tab:imagenet_1k}
    \vspace{-5pt}
\end{table}

\begin{table}[]
    \small
    \centering
    \begin{tabular}{lcccccc}
     \HHLINE 
     \textbf{Method} & \textbf{Train Iters} & \textbf{UNet} & \textbf{FID} & \textbf{IS} \\
     \hline
     SiT-XL/2 (Baseline) & 200k & False & 25.2 & - \\
     SiT-XL/2 ($\alpha_{\rm min}$ = 1.0)& 200k & True & 22.7  & 57.3 \\
     SiT-XL/2 ($\alpha_{\rm min}$ = 0.8) & 200k & True & \textbf{14.2} & \textbf{79.7} \\
     \hline
     SiT-XL/2 (Baseline) & 400k & False & 17.2 & - \\
     SiT-XL/2 ($\alpha_{\rm min}$ = 1.0)& 400k & True & 16.5  & 74.8 \\
     SiT-XL/2 ($\alpha_{\rm min}$ = 0.8) & 400k & True & \textbf{11.8} & \textbf{91.8} \\
    \end{tabular}
    \caption{\textbf{Class-conditional ImageNet-1k 256x256 Generation with SiT-XL/2~\cite{ma2024sit}.} 
    Our method significantly enhances the generation performance and the convergence speed. Comparison between ours ($\alpha_{\rm min}$ = 0.8) and UNet baseline ($\alpha_{\rm min}$ = 1.0) suggests the improvements are primarily from our decayed identity shortcuts rather than the encoder-decoder skip connections.}
    
    \label{tab:sit_xl}
\end{table}
\begin{table*}[th!]
\small
    \centering
    \scalebox{0.9}{
    \begin{tabular}{cc|ccccc|cccc}
        \HHLINE
        \shortstack{Feat.\\Dim.}&\shortstack{Enc. \\Depth } ($L$) & \multicolumn{5}{c|}{$\alpha_{\rm min}$} & \multicolumn{4}{c}{$\alpha_{L}^{\rm eff}$}\\
         \hline
         & & 0.6&  0.7& 0.8 & 0.9 & 1.0 (Baseline) & (0, 1e-3) & [1e-3, 1e-2) & [1e-2, 1e-1) & [1e-1, 1]]\\
        \hline 
        384 & 12 & \textbf{78.5} & 78.1 & 75.2 & 73.5 & 69.2 & - & \textbf{78.5} & 78.1 & 75.2 \\
        768 & 12 & \textbf{83.6} & 81.8 & 79.8 & 79.2& 76.5 & - & \textbf{83.6} & 81.8 & 79.8\\
        1024 & 12 & \textbf{83.2} & 82.5 & 82.1 & 79.3& 78.0 & - & \textbf{83.2} & 82.5 & 82.1 \\
        768 & 18 & 83.5 & \textbf{85.0} & 84.4 & 81.8& 79.2 & 78.5 & \textbf{85.0} & 84.4 & 81.8\\
        1024 & 24 & 84.3 & \textbf{86.0} & 84.5 & 84.3& 81.4 & 82.4 & \textbf{86.0} & - & 84.3\\
        \HHLINE
        \end{tabular}}
        \vspace{-5pt}
        \caption{\textbf{Linear Probing accuracy of MAE on ImageNet-100 varying $\alpha_{\rm min}$ and architecture.} We show our method consistently improves the performance across all configurations. We notice the encoder depth rather than feature dimension influences the optimal $\alpha_{\rm min}$. We attribute this behavior to the scaling effect of the input data to encoder's final layer, quantified as $\alpha_{L}^{\rm eff}=\prod_{l=1}^L\alpha_l$. Deeper model requires larger $\alpha_{\rm min}$ to maintain a consistent cumulative decay effect and we find setting $\alpha_{\rm min}$ such that $\alpha_L^{\rm eff}\in$ [1e-3, 1e-2) works the best. With our strategy, smaller model (768 feat. dim + 12 layer) outperforms bigger one (1024 feat. dim + 24 layer) with standard residual connections.}
        \label{tab:ablation_imagenet_100}
\end{table*}
\begin{table*}[th!]
\small
\centering
    
\begin{subtable}[t]{0.45\linewidth}{
\centering
\hspace{40pt}\begin{tabular}[t]{c|c|c|c}
\HHLINE
\centering
\textbf{$\alpha_{\rm min}$} & \textbf{$\alpha_l$ scheduler} & \textbf{UNet} & \textbf{LP} \Tstrut\Bstrut \\
\hline
$0.6$ & Linear & Yes & \bf 83.6 \\
\hline
$0.6$ & Linear & No  & 61.5 \\
$0.6$ & Cosine & Yes & 82.8 \\
$0.7$ & Linear & Yes & 81.8 \\
$0.7$ & Cosine & Yes & 82.9 \\
-  & Learnable $\alpha_l$ & Yes & 79.5 \\
\HHLINE
\end{tabular}
}\caption{\textbf{Effect of Skip Connections and $\alpha_l$ scheduler}. Skip connection is critical for performance. The choice of schedulers has less impact and learnable $\alpha_l$ is worse than pre-fixed $\alpha_l$.
\label{tab:skip_connections}}
\end{subtable}%
\hfill%
\begin{subtable}[t]{0.52\linewidth}{
\centering
\scalebox{1.0}{
\begin{tabular}[t]{l|c|c|c}
\HHLINE
\small
\textbf{Configurations} & \textbf{Decay Block} & $\alpha_{\rm{min}}$ & \textbf{LP} \Tstrut\Bstrut \\
\hline
$\vx_{l+1} = \vx_l + f_{\btheta_l}(\vx_l)$ & - &  $-$ & 76.5\\
$\vx_{l+1} = \vx_l + \sqrt{0.5} f_{\btheta_l}(\vx_l)$ & MLP \& Atten. &  $-$ & 76.9\\
$\vx_{l+1} = \sqrt{0.5} \left(\vx_l + f_{\btheta_l}(\vx_l)\right)$ & MLP \& Atten. & $-$ & 82.6\\
$\vx_{l+1} = \alpha_l\vx_l + f_\theta(\vx_l)$ & Atten. & 0.6 & 79.3 \\
$\vx_{l+1} = \alpha_l\vx_l + f_\theta(\vx_l)$& MLP & 0.6 & 80.6 \\
$\vx_{l+1} = \alpha_l\vx_l + f_\theta(\vx_l)$ & MLP \& Atten. & 0.6 & \textbf{83.6}\\
\HHLINE
\end{tabular}}
\caption{\textbf{Other Decay Schemas}. We conduct ablations using a variety of scalings of the residual connection and observe that our design produces the best results.
\label{tab:decay_schema}}
}\end{subtable}
\vspace{-5pt}
\caption{\textbf{Ablation experiments of MAE using ViT-B/16 in ImageNet-100.} We ablate decay scheduler and architectural design with linear probing (LP) accuracy.}
\label{tab:extra_ablation}
\end{table*}

We report the results in Table \ref{tab:imagenet_1k}. 
In the top half of the table, we present methods that employ a contrastive loss. Although these methods produce the best probing accuracies, their success depends on a carefully designed data augmentation process, 
which may need to be tuned for each different data distribution. 
In the bottom half, we show several methods based on generative architecture. 
Our method simply extends MAE by constructing an implicit feature bottleneck and shows significant improvements over the MAE baselines for both linear probing (72.7\% \emph{vs.} 67.3\%) and K-Nearest Neighbour (63.9\% \emph{vs.} 27.4\%). 
outperforming Data2Vec, Latent MIM and CAE and giving a probing accuracy competitive with I-JEPA, without needing to use explicit feature alignment.

End-to-end fine-tuning (FT), unlike linear probing which only trains a single linear layer, updates the entire network for image classification. Since the features can shift significantly from their pre-training state during end-to-end updating, we argue that this may not accurately reflect the quality of the learned representations. For example, DINO demonstrates superior performance in various downstream vision tasks compared to MAE, but its fine-tuning performance is worse than MAE. Similarly, ConMIM and ADDP exhibit poor linear probing performance, suggesting lower-quality representations, yet their fine-tuning performance surpasses that of contrastive learning methods. Nevertheless, we still provide the fine-tuning results for reference.

\subsection{Ablation Studies on ImageNet-100}
\label{sec:ablation}
We conduct ablations on several properties of our framework on ImageNet-100. A summary of results can be found in Tables~\ref{tab:ablation_imagenet_100} and~\ref{tab:extra_ablation}.

\noindent
\textbf{Decay rate $\alpha_{\rm min}$.} 
The only parameter of our framework is $\alpha_{\rm min}$, the minimum scaling factor applied to the final layer. In Table~\ref{tab:ablation_imagenet_100},we show linear probing scores for varying values of $\alpha_{\rm min}$. We observe that $\alpha_{\rm min} \in [0.6, 0.7]$ works well for most cases. If $\alpha_{\rm min}$ is too small, for example, $\alpha_{\rm min} \leq 0.4$, we observe that the training becomes unstable.

\noindent
\textbf{Architecture size.} In Table~\ref{tab:ablation_imagenet_100}, we also run experiments over multiple architecture choices. We notice encoder depth $L$ rather than feature dimension that influences the optimal choices of $\alpha_{\rm \min}$ and deeper models require larger $\alpha_{\rm \min}$. We attribute this phenomenon to the cumulative decaying effect of the final layer, quantified as $\alpha_{L}^{\rm eff} = \prod_{l=1}^L\alpha_l$. Heavy decaying would harm the optimization and selecting $\alpha_{\rm \min}$ such that $\alpha_{L}^{\rm eff}\in$ [1e-3, 1e-2) yields the best performance. 


\noindent
\textbf{Skip connections.} Another critical design choice in our network is to include skip connections that are not in the original MAE. As discussed in Section~\ref{sec:method_impl}, if the MAE does not use skip connections, the bottleneck layer must preserve all information to reconstruct the input image accurately. This is opposed to learn abstract representations at bottleneck. These contrary effects significantly degrade the representation learned by the model, leading to a 22.1\% drop in the linear probing score, as we report in Table~\ref{tab:skip_connections}. 

\noindent
\textbf{$\alpha_{l}$ Scheduler.} We use linear scheduler $\alpha_l = 1-\frac{(1-\alpha_{\rm min})}{L}l$ as a default choice. In table~\ref{tab:skip_connections}, we also experiment with a cosine scheduler but find it leads to worse performance for $\alpha_{\rm \min} = 0.6, 0.7$. Besides using a prefixed $\alpha_l$ scheduler, we also experiment with a learnable $\alpha_l$, which resembles the setup of highway network \cite{srivastava2015highway}. We show the learnable $\alpha_l$ at each layer in Table\ref{tbl:learn_alpha}, appendix. From the table, we don't find consistent patterns over network depth and the performance is worse than our predefined $\alpha_l$ scheduler. 

\noindent
\textbf{Different decay schema.}
We also explore decay schema, with results summarized in Table~\ref{tab:decay_schema}: (1) Scaling both branches of the residual blocks simultaneously by applying a constant factor, $\alpha = \sqrt{0.5}$, to both $\vx$ and $f_{\btheta_l}(\vx)$. (2) Scaling only $f_{\btheta_l}$ using the same constant factor, $\alpha = \sqrt{0.5}$. (3) Applying our proposed schema exclusively to either the attention or MLP branch.

Among these, (2) shows no significant improvement over the baseline, while (1) yields some improvement but still underperforms compared to our approach. By analyzing (1) and (2), we demonstrate that the representation gains are due to down-weighting the skip connection branch. Notably, recent diffusion models \citep{karras2017progressive, song2020score} have employed (1) in their smaller convolutional neural network but don't provide systemic analysis. However, applying decay only to the MLP or attention branch reduces the overall decaying effect across the network, resulting in lower performance compared to our schema, which achieves the best performance among the tested designs.

\begin{figure}
\centering
\begin{minipage}{1.0\linewidth}
\centering
        \begin{subfigure}[t]{1.0\textwidth}
        \includegraphics[width =1.0\textwidth]{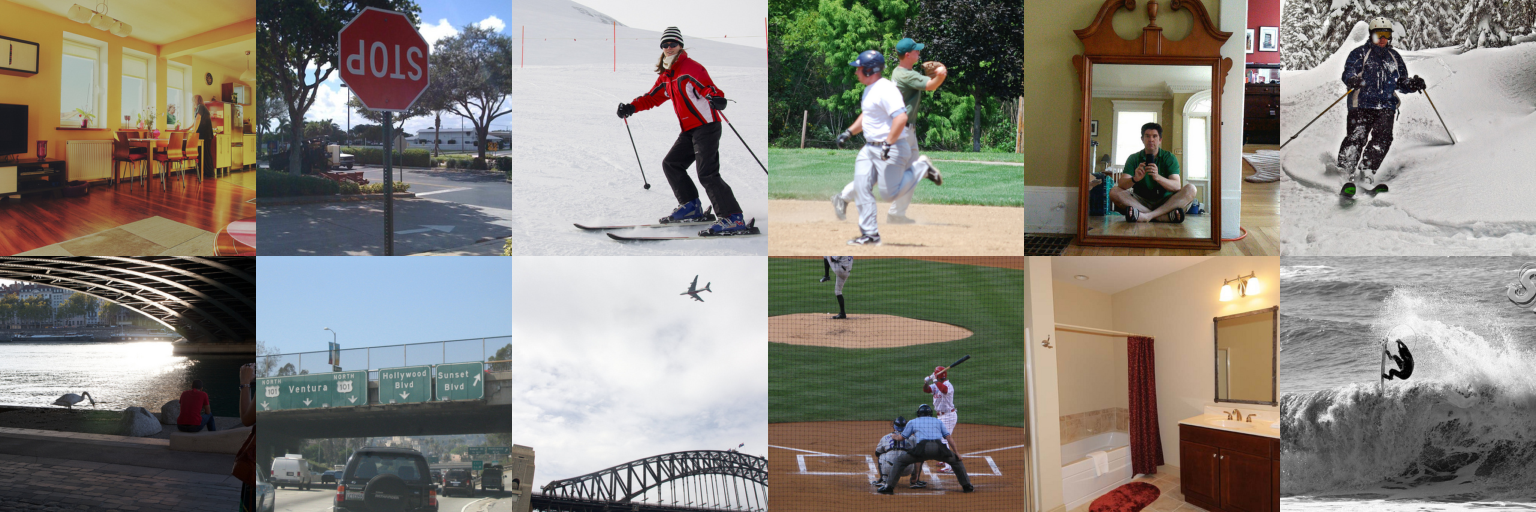}
        \caption{Input Image}
        \end{subfigure}\hfill
        \begin{subfigure}[t]{1.0\textwidth}
         \includegraphics[width =1.0\textwidth]{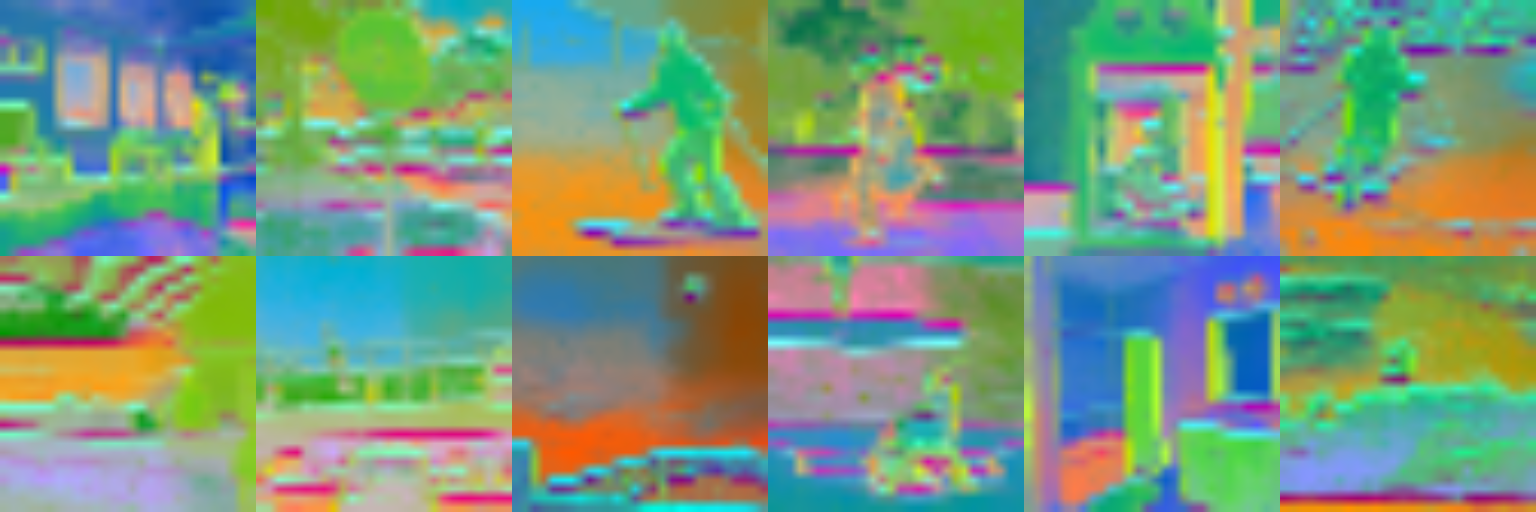}
        \caption{MAE {{(4.1 mIoU)}}}
        \end{subfigure}\hfill
        \begin{subfigure}[t]{1.0\textwidth}
        \includegraphics[width =1.0\textwidth]{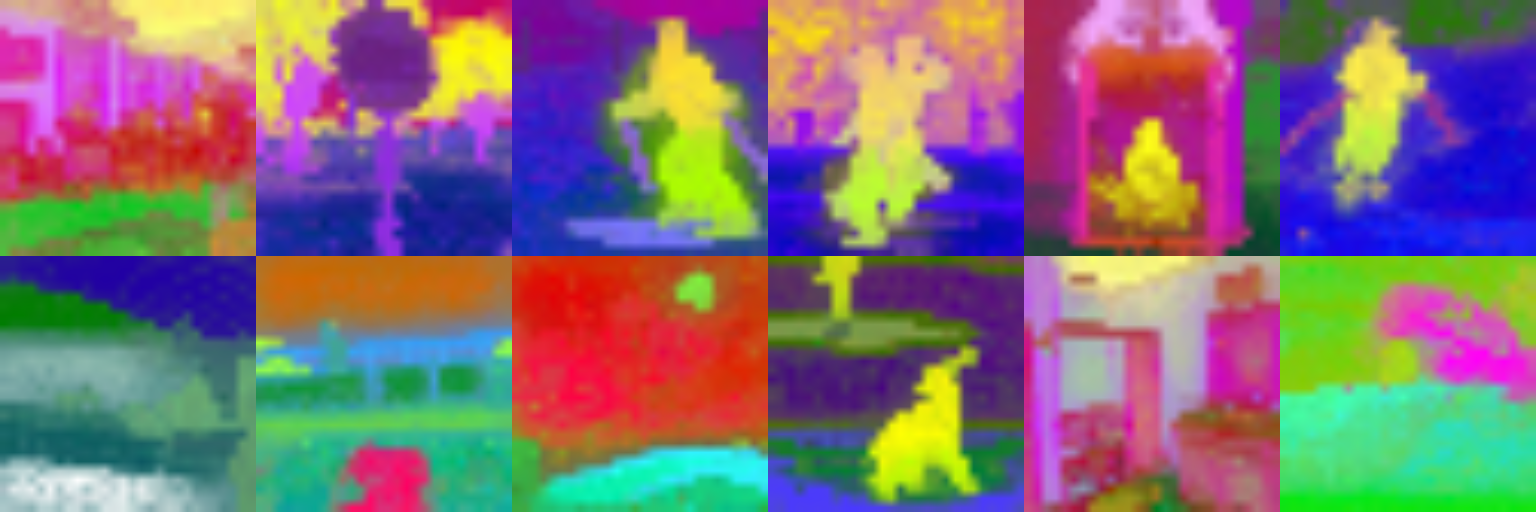}
        \caption{MAE ($\alpha_{\rm min}$ = 0.6) {(10.4 mIoU)}}
        \end{subfigure}
\caption{\textbf{Visualize learned representations using \citet{zhang2024deciphering} without cherry-picking.} 
We project the learned representations onto a 3-channel feature map, visualized as RGB images. Our method learns more abstract and semantically consistent representations. This visual comparison is further supported by benchmarking on unsupervised semantic segmentation tasks, where our approach achieves better results (10.4 mIoU) compared to the baseline MAE (4.1 mIoU) on COCO validation set.}
\label{fig:vis_eigen}
\end{minipage}
\vspace{-7pt}
\end{figure}

\subsection{Embedding Analysis} 
We qualitatively evaluate the feature learning in Figure~\ref{fig:vis_eigen} and we adopt the pixel-wise embedding approaches proposed by~\citet{zhang2024deciphering} to group the 
representations from the last layer of the encoder into a lower dimensional space. We use their default hyperparameters to cluster representations across COCO validation set and render the top 3 eigenvectors as RGB channels of images. From the visualization, ours learns abstract representation and the object from the same categories have similar color, indicating a global consistent semantic grouping. The baseline MAE, on the other hand,  doesn't show clearly global semantic patterns.

We also benchmark the clustering quantitatively, following the postprocessing protocol ~\citep{zhang2024deciphering} to produced unsupervised semantic segmentation and report the results as the mean intersection of union (mIoU). Ours (10.41 mIoU) achieve 6.31 mIoU improvement over baseline (4.10 mIoU), which support the visual comparison.
\vspace{-5pt}
\section{Experiments on Diffusion Models}

\textbf{Diffusion models.}
We use U-ViT~\citep{bao2023all}, a ViT-based diffusion model with skip connections between the encoder and decoder, and SiT-XL/2~\cite{ma2024sit} as the baseline for our diffusion model experiments. To adapt SiT-XL/2 to our design, we add skip connection from the first half of the layers to the corresponding second half of the layers. 
Unlike in MAE where we only apply decayed connections within the encoder, in diffusion models, we decay the skip connections till the last layer of the decoder, following the recent studies \citep{yang2023diffusion, baranchuk2021label} suggesting that diffusion models learn the best semantic representations near the decoder's latter stages. While this design might be suboptimal, as the smallest decay factor may not align with the layers holding the best semantic representations, we demonstrate in practice that this simple approach effectively enhances both the learned representations and the quality of generated outputs.

\noindent
\textbf{Experimental details.}
We utilize the default scheduler and sampler from U-ViT~\citep{bao2023all} and SiT-XL/2~\cite{ma2024sit} respectively. We train class-conditional diffusion model on ImageNet-100 and ImageNet-1k and we additionally train unconditional diffusion models on CIFAR-100 and ImageNet-100 without using image class labels to validate our design across tasks. For ImageNet-100 and ImageNet-1k, we follow the setup in latent diffusion models ~\citep{rombach2022high} by running the model in the latent space of a pretrained VAE, which maps input images with 256x256x3 to a 32x32x4 sampled latent space. In ablation experiments with U-ViT, We use U-ViT-Mid for ImageNet-100 and ImageNet-1K, and U-ViT-small for CIFAR-100. For model design and training details, please refer to ~\citet{bao2023all}.

We evaluate the learned representations with linear probing and we train a linear classifier over the frozen representations. We report the results as the best configurations, including the choices of layer index and noise level, that yields the best performance. 

\noindent
\textbf{Results.} We report our primary ImageNet-1k 256x256 class-conditional image generation performance with SiT-XL/2 in Table~\ref{tab:sit_xl}. Compared to baseline, we demonstrate significant improvement in FID: (11.8 \emph{v.s.} 17.2). Further, comparing ours with SiT-XL/2 ($\alpha_{\rm min}$ = 1.0) (11.8 \emph{v.s.} 16.5) further validates that our improvements are primarily from our decay scheme rather than the encoder-deocder skip connection. Our simple design also shows significant convergence speed up where ours at 200K steps are substantially better than baseline at 400 steps (14.2 \emph{v.s.} 17.2).

Besides, we also provide ablation with U-ViT models across multiple dataset and experimental setup. We present our results in Table~\ref{tbl:diffusion}, where we demonstrate that replacing residual connections with our proposed decayed identity shortcuts consistently enhances representation quality and image generation across both datasets and tasks (conditional and unconditional generation). Notably, this improvement is achieved without introducing any additional learnable parameters. We provide the visualization of generated images in the appendix for qualitative comparisons (Figure \ref{fig:diffusion_images}).

\begin{table*}[t]
\small
    \centering
    \begin{tabular}{c|c|cccc|cccc}
    \HHLINE
    \multicolumn{1}{c|}{} & Model & \multicolumn{4}{c|}{\textbf{Linear Probing (Acc.)}$\uparrow$} & \multicolumn{4}{c}{\textbf{Generation quality (FID)}$\downarrow$} \\ 
    \hline
    \backslashbox{Dataset}{$\alpha_{\rm min}$} & &  1.0 & 0.8 & 0.7 & 0.6 & 1.0 & 0.8 & 0.7 & 0.6 \\
    \hline
    CIFAR-100 (Uncon.) & U-ViT-Small & 62.47 & 63.58 & \textbf{66.86} & 64.63 &  14.34 & 11.65 & \textbf{8.99} & 11.71\\
    ImageNet-100 (Uncond.) & U-ViT-Mid & 72.8 & 74.5 & \textbf{76.1} &  75.8 &  44.40 & \textbf{40.96} & 41.17 & 43.51\\
    ImageNet-100 (Class Cond.) & U-ViT-Mid & - & - & - & -&  6.93 & 5.75 & 5.11 & \textbf{4.98}\\
    ImageNet-1K (Class Cond.) & U-ViT-Mid & - & - & - & -&  7.65  & 6.23 & 5.34 & \textbf{4.90}\\
    \HHLINE
    \end{tabular}
    \caption{ \textbf{Ablation on image generation using U-ViT models.}
    In diffusion models, we demonstrate that our proposed decayed identity shortcut (with $\alpha_{\rm min} < 1.0$) enhances probing accuracy and generation quality across various configurations in both classes conditional and unconditional setups.}
    \label{tbl:diffusion}
\end{table*}

\section{Discussion on Feature Rank}
\label{sec:discussion}
\newcommand{\mcr}{MCR$^2$ }
In this section, we try to provide insights for the key question: 
How and why do residual connections impact the abstraction level of the deeper layers in a neural network? We delve deeper into how our design reinforces the low-rank bias of neural networks and try to connect our method to ideas in existing works \citep{huh2021low}. To this end, we visualize the training dynamics of our method and analysis the feature rank of our approach to provide a holistic analysis.

\noindent
\textbf{Low-rank simplicity bias.} \citet{huh2021low} investigate the low-rank simplicity bias in deeper feed-forward neural networks, which drives neural networks to find low-rank solutions. 
At the same time, they make an empirical observation that deeper residual networks do not show a similar rank contracting behavior.

\noindent
\textbf{Effective rank.}
For analysis purpose, \citet{huh2021low} quantify the rank of the learned representation using the \textit{effective rank}, which is a continuous measure.
For a matrix $\mA\in \mathbb{R}^{m\times n}$, the effective rank $\rho(\mA)$ is defined as the Shannon entropy of the normalized singular values \citep{roy2007effective}:
$\rho(\mA) = -\sum_{i}^{\min(n,m)}\bar{\sigma}_i\log \bar{\sigma}_i$
where $\bar{\sigma}_i = \sigma_i / \sum_{j} \sigma_j$  denotes the $i^\text{th}$ normalized singular value. Intuitively, $\rho(\mA)$ is small when a few singular values dominate and large when singular values are evenly spread, hence giving a good continuous approximation for matrix rank.
In the following subsections, 
we use the singular values from the covariance matrix $\mA_\btheta$ of the last-layer features to compute $\rho(\mA)$, where $\mA_\btheta(i,j)$ denotes the covariance of the learned class tokens for the $i^\text{th}$ and $j^\text{th}$ samples.

\label{sec:unsupervised}
\begin{figure}[t]
    \centering
    \begin{minipage}{1.0\linewidth}
    \begin{minipage}{1.0\linewidth}
    \begin{subfigure}{1.0\linewidth}
        \centering
        \frame{\includegraphics[width=0.6\textwidth]{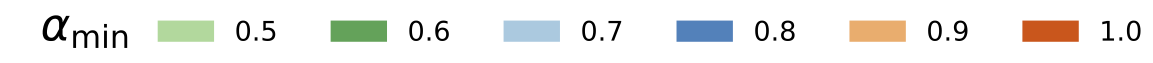}}
    \end{subfigure}
    \end{minipage}
    \begin{minipage}{.49\linewidth}
    \begin{subfigure}{1.0\linewidth}
        \centering
        \includegraphics[width=1.0\textwidth]{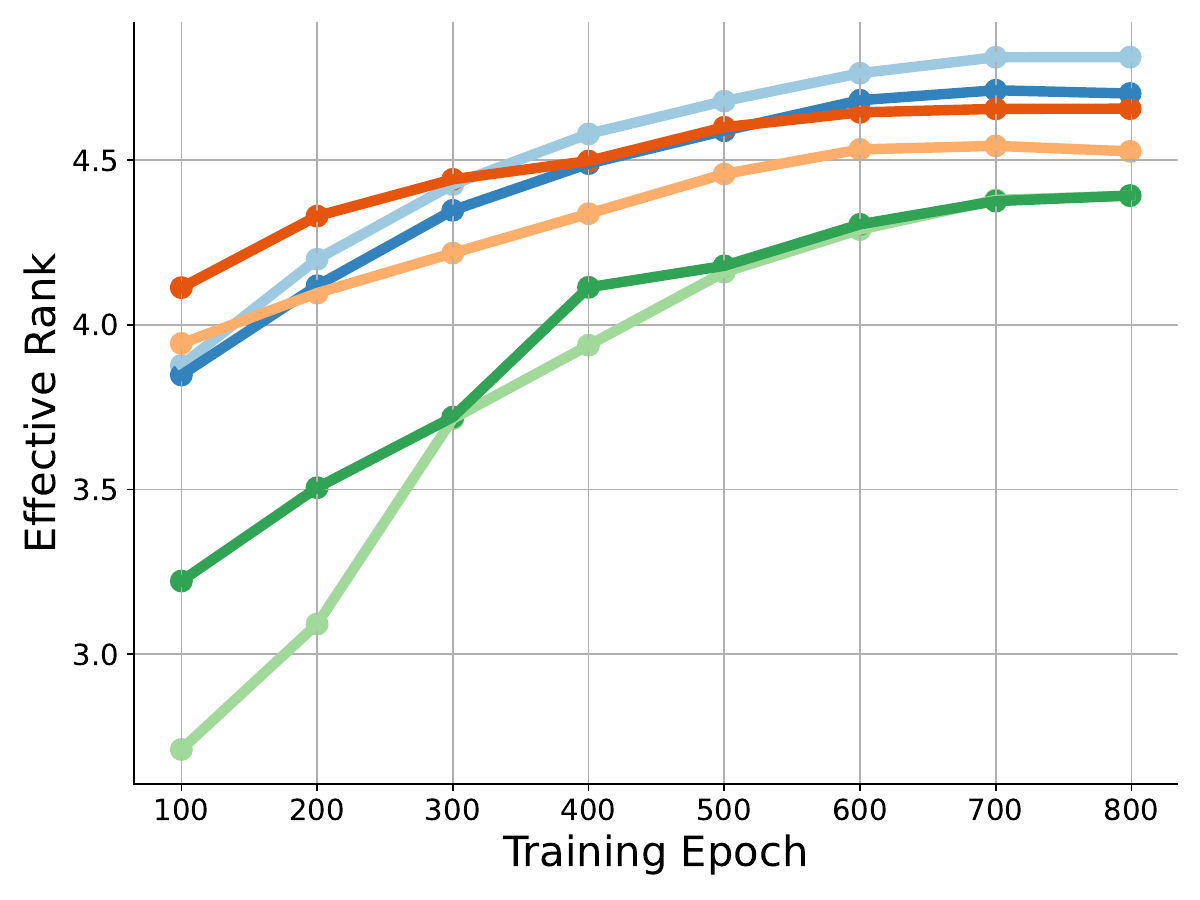}
        \caption{Effective Rank of MAE over training epochs.}
        \label{fig:effective_rank_unsupervised}
    \end{subfigure}
    \end{minipage}%
    \hfill%
    \begin{minipage}{.49\linewidth}
    \begin{subfigure}
    {1.0\linewidth}
        \centering
        \includegraphics[width=1.0\textwidth]{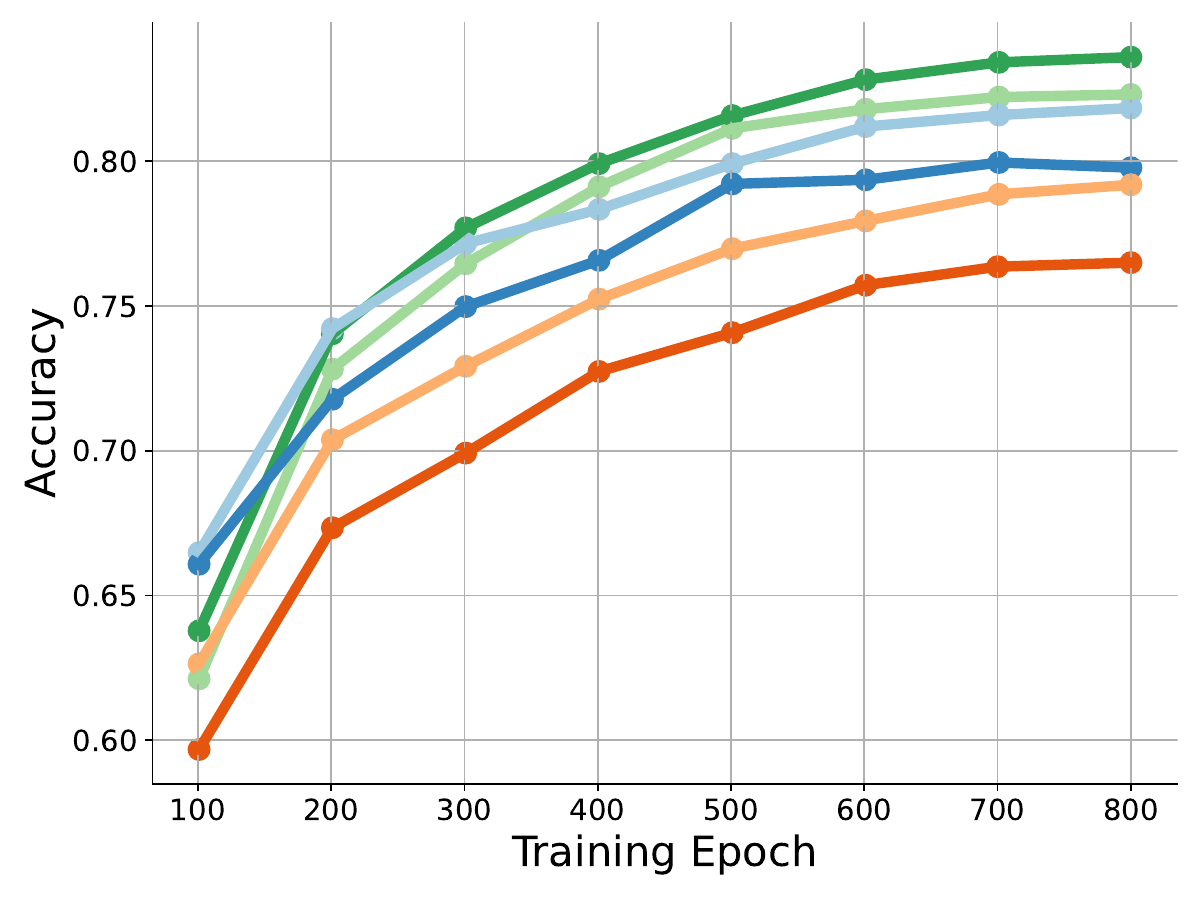}
        \caption{Linear probing accuracy of MAE over training epochs.}
        \label{fig:acc_unsupervised}
    \end{subfigure}
    \end{minipage}
    \end{minipage}%
    \hfill%
    \caption{
For MAE pretrained on ImageNet-100, we present visualizations of (a) the training dynamics of the effective rank for different values of $\alpha_{\rm{min}}$, and (b) the linear probing accuracy for various $\alpha_{\rm{min}}$, demonstrating that a lower effective feature rank could potentially be associated with better performance.}
    \label{fig:unsupervised}
\vspace{-5pt}
\end{figure}

Inspired by \citet{huh2021low}, \textit{we conjecture that the feature learning improvement of our method can potentially be attributed to the decayed identity shortcuts encouraging low-rank features at the bottleneck.}
In Figures~\ref{fig:effective_rank_unsupervised} and~\ref{fig:acc_unsupervised}, we measure the training dynamics of the models (feature dimension 768 and encoder depth 12) in terms of accuracy and the effective rank, for different values of $\alpha_{\text{min}}$.
In early epochs, models with lower $\alpha_{\text{min}}$ tend to exhibit both lower effective rank and higher probing accuracy, supporting our hypothesis. As training progresses, the correlations between $\alpha_{\text{min}}$ and effective rank become less precise. We suspect this is due to the model's effort to compensate for the decay factors with a large value. And we can still see that lower $\alpha_{\min} =$[0.5-0.6] results in lower feature rank and better probing accuracy compared to $\alpha_{\min}$ = [0.7-1.0]. We provide further analysis in Appendix~\ref{app:theoretical}.

\noindent
\textbf{Compatibility with contrastive learning frameworks.}
\label{paragraph:limitation} Despite substantial improvement of applying our decayed identity shortcuts in generative models, we note that our approach does not easily extend to contrastive learning frameworks, where the low rank inductive bias conflicts with the training objectives, e.g., MoCov3~\citep{chenempirical} include a universal repulsive term in the denominator to increase the feature rank. Rankme~\citep{garrido2023rankme} confirms this by showing contrastive learning models prefer higher feature ranks. 

\section{Conclusion}
\citet{huh2021low} raise a key insight in their work -- that how a neural network is parameterized matters for fitting the data -- and investigate the inductive low-rank bias of stacking more linear layers in a network.
In this work, we observe that the ubiquitous residual network~\citep{he2016deep}  may not be the ideal network parametrization for representation learning and propose a modification of the shortcut path in residual blocks that significantly improves unsupervised representation learning. We explore the connection between our reparameterization of the residual connection and the effective rank of the learned features, finding a potential correlation between good representations and low-rank representations.

Our work calls into question a fundamental design choice of neural networks that has been used in many modern architectures. By rethinking this choice, the door is open for further reparametrizations and improvements to unsupervised representation learning. 
The results we show provide a prompt for more extensive investigations into the connection between low effective rank and high-quality abstract representations.

\section{Acknowledgments}
We gratefully acknowledge the support of AFOSR FA9550-18-1-0166, NSF DMS-2023109, DOE DE-SC0022232, the NSF-Simons AI-Institute for the Sky (SkAI) via grants NSF AST-2421845 and Simons Foundation MPS-AI-00010513, and the Margot and Tom Pritzker Science Foundation.

{
    \small
    \bibliographystyle{ieeenat_fullname}
    \bibliography{cvpr2026}
}
\newpage
\appendix
\section{Training and Evaluation Details}
\label{sec:appendix_experiments_details}
\subsection{Model training.} 
Our training configurations primarily followed the guidelines established by \citet{he2022masked}. 
In the ImageNet-1K experiment, our model was trained for 800 epochs, utilizing the AdamW~\cite{loshchilov2017decoupled} optimizer with a constant weight decay of 5e-2 for a batch size of 1024.
We set the maximum learning rate to 6e-4. Initially, the learning rate started at 0 and linearly increased to its maximum over the first 40 epochs, after which it followed a cosine schedule to gradually decrease to zero by the end of the training period. 
It is worth noting that the learning rate per sample, or effective learning rate, in our setup matched that of \citet{he2022masked}, although our maximum learning rate was set lower due to our batch size being a quarter of theirs.
We applied random resizing, cropping, and horizontal flipping during training as part of our augmentation scheme.
To enhance the quality of the learned representations in most experiments, we employed the normalized pixel loss, as proposed by \cite{he2022masked}.
In the ImageNet-100 experiment, we employed the identical training configuration used in the ImageNet-1K experiments. We train our model with 4 NVIDIA A40 GPUs and a completed trianing usually takes 20 hours on ImageNet-100 and 200 hours on ImageNet-1k. 

\subsection{Evaluation with Linear Probing.}
For the ImageNet-1k dataset, we use the exact same evaluation protocols employed in \citet{he2022masked}, which includes random data augmentation. 

For the ImageNet-100 dataset, we employed a simpler evaluation protocol: We train the linear classifier with a batch size of 1024 for 200 epochs, where the learning rate starts at 1e-2 and then decays towards 0 using a cosine scheduler. During this evaluation, we do not apply any data augmentation.

\begin{figure}[!htbp]
   \centering
   \includegraphics[width=.89\linewidth]{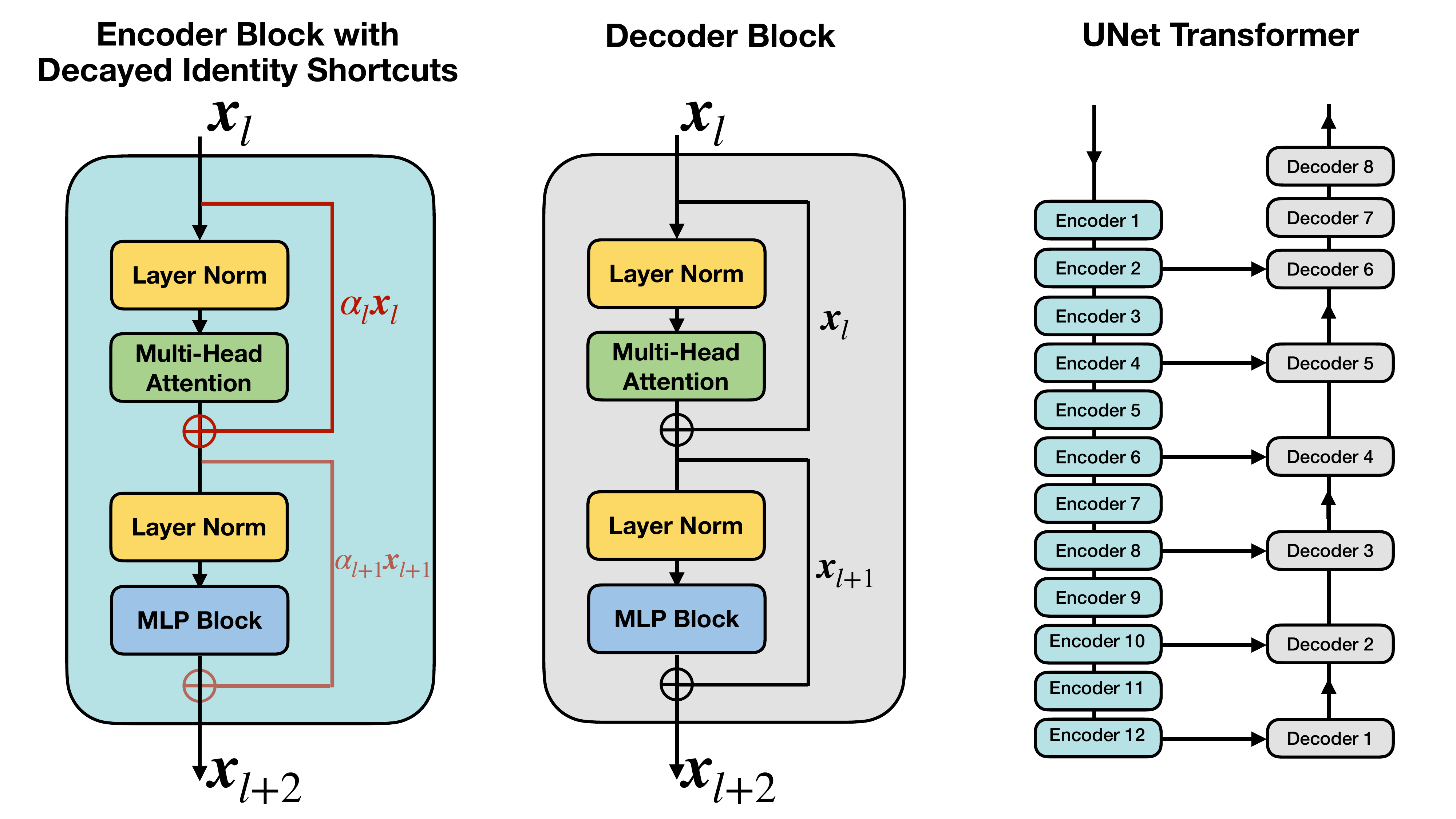}
   \caption{We present our enhanced UNet Transformer architecture for Masked Auto-encoder. 
   (1) \textbf{\textit{Left}}: Our customized encoder blocks, equipped with our proposed decay identity shortcuts.
   (2) \textbf{\textit{Middle}}: Standard transformer blocks as the decoder blocks.
   (3) \textbf{\textit{Right}}: We incorporate the decay identity shortcuts exclusively within the encoder blocks of our UNet transformer and employ standard transformer blocks for the decoder. 
   To support abstract representation learning at the bottleneck, \emph{i.e.,} the last layer of the Encoder 12, we adopt the UNet~\cite{ronneberger2015u} architecture and create skip connections that transmit every other encoder feature directly to the decoder.}
   
   \label{fig:appendix}
\end{figure}

\subsection{Modified Architecture}
We present a visualization of our UNet transformer design, as outlined in Section ~\ref{sec:method_impl}, in Fig.~\ref{fig:appendix}. It's important to note that decayed identity shortcuts are exclusively implemented within the encoder block. Additionally, we establish skip connections from alternating blocks in the encoder to the decoder, following the UNet~\cite{ronneberger2015u} architecture's design principles.

\subsection{Learnable $\alpha_l$ over network layers}
\begin{table*}[t]
\small
    \centering
        \begin{tabular}{lcccccccccccc}
        \hline
        Layer Index & 	1	& 2	& 3	& 4	& 5&6&	7&	8&	9&	10&	11&	12\\
        \hline
        Attention & 0.993 & 	0.947&	0.982&	0.766&	0.992& 	0.795&	0.988&	0.849&	0.998&	0.723&	0.811&	0.488\\
        FFN  &	0.989	& 0.926&	0.961&	0.620&	0.961&	0.442&	0.711&	0.322&	0.810&	0.475&	0.637&	0.353
        \end{tabular}
        \caption{Learnable $\alpha_l$ values for different model layers. In contrast to our proposed linear scheduler, learnable $\alpha_l$ does not exhibit a consistent decay pattern across network depth.}
        \label{tbl:learn_alpha}
\end{table*}
        
In the ablation study of learnable $\alpha_l$, we apply no additional regularization beyond standard weight decay to the model parameters. Table \ref{tbl:learn_alpha} presents the $\alpha_l$ values for each layer. The results do not reveal any meaningful pattern across network depth.

\section{Further analysis}
\subsection{Reconstruction quality.}
\begin{figure}[ht!]
    \centering
    \includegraphics[width=0.5\textwidth]{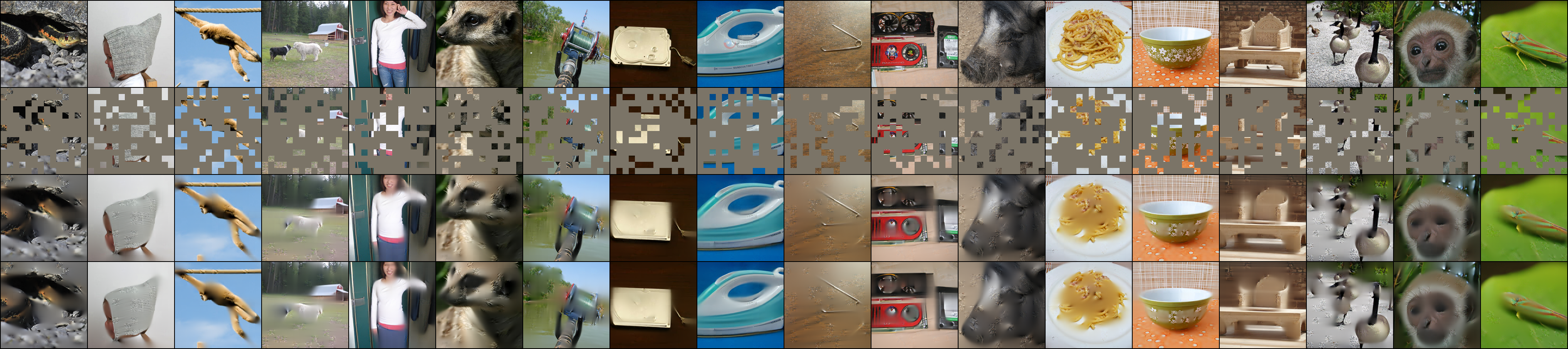}
    \caption{\textbf{Qualitative comparison of images reconstructed by MAE with and without our method.} We observe our method learns features with higher linear probing accuracy without compromising reconstruction quality. Row 1: ground truth test image. Row 2: images masked at $75$\%. Row 3: reconstructions with our method. Row 4: reconstructions with baseline MAE.}
    \label{fig:recon}
\end{figure}
We qualitatively evaluate test images reconstructed by an MAE using our framework and images reconstructed by the original MAE. We show the reconstructed images in Figure~\ref{fig:recon}. While the focus of our work is entirely to improve the representations learned by an encoder, we observe that our framework does not harm the reconstructions. Hence, there is no qualitative tradeoff for our increase in linear probing accuracy.

\subsection{Abstraction and Low-rank in the Supervised Setting}
\begin{figure}[ht!]
    \centering
    \begin{subfigure}[t]{1.0\linewidth}
        \centering
        \frame{\includegraphics[width=0.6\textwidth]{images/legend.png}}
    \end{subfigure}
    \vspace{-15pt}
    \begin{center}    
    \begin{subfigure}[t]{.40\linewidth}
        \hspace{-20pt}
        \centering
        \includegraphics[width=1.0\textwidth]{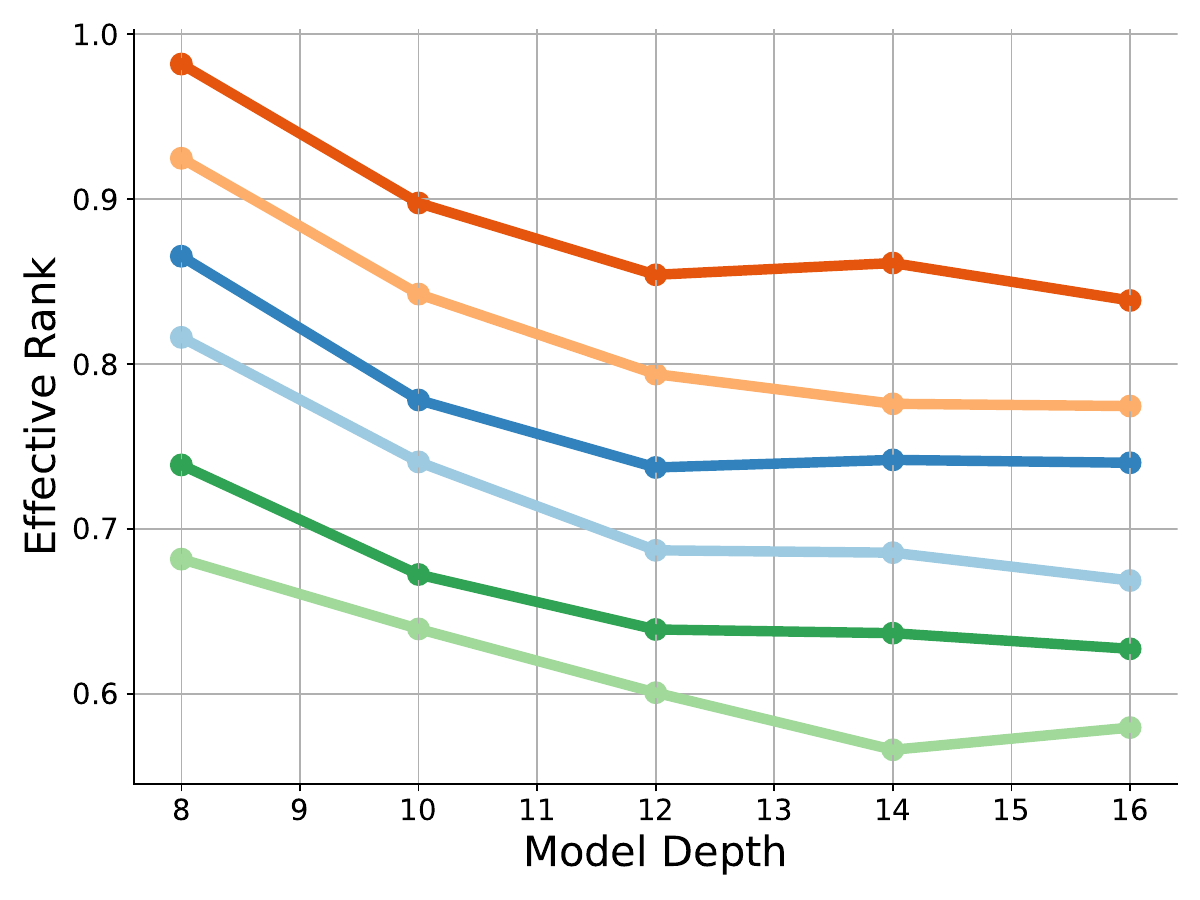}
        \caption{Effective rank of ResNet for different depths at the convergence of the training.}
        \label{fig:effective_rank_supervised_depth}
    \end{subfigure}
    \hspace*{10pt}
    \begin{subfigure}[t]{.40\linewidth}
        \centering
        \includegraphics[width=1.0\textwidth]{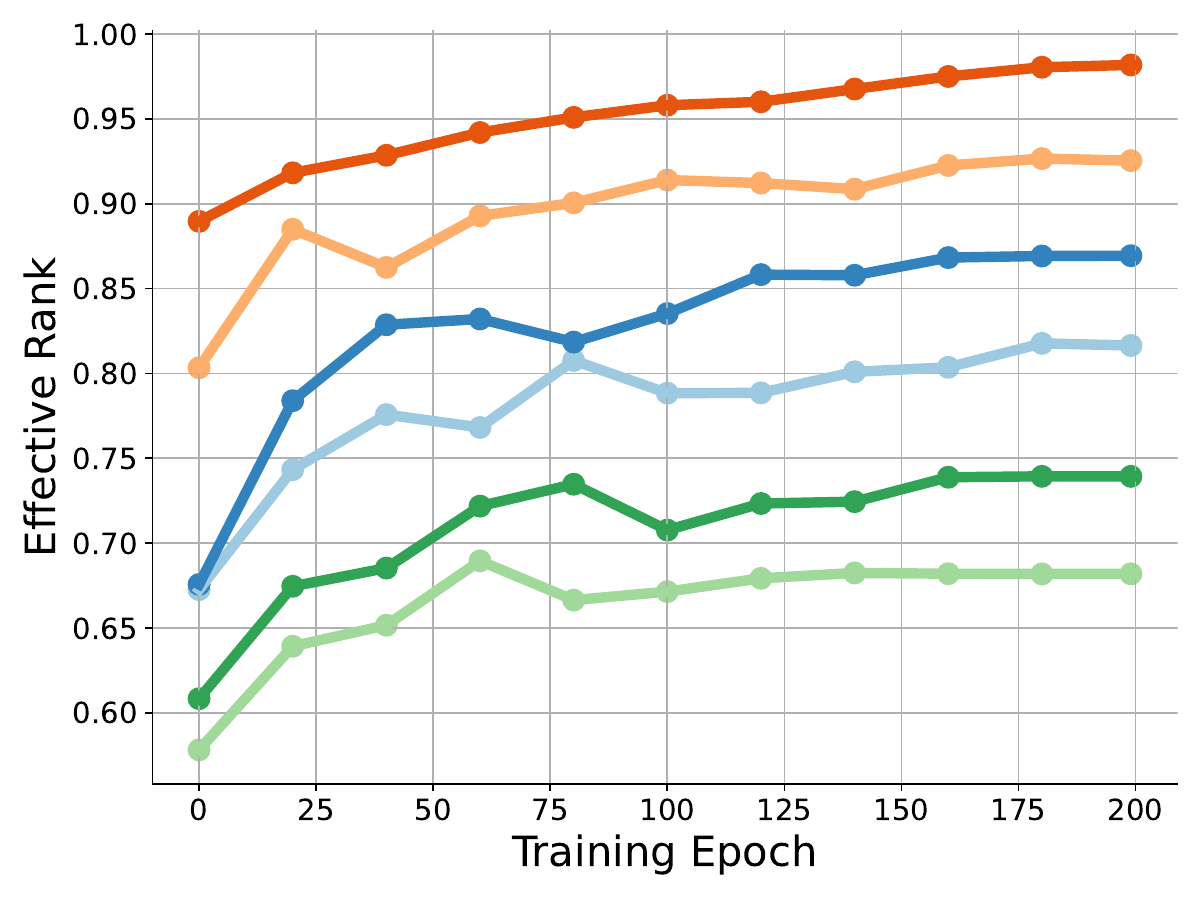}
        \caption{Effective rank of ResNet over training epoch.}
        \label{fig:effective_rank_supervised_time}
    \end{subfigure}
    \end{center}
    \caption{\textbf{Dynamics of the feature rank in the supervised setup.}
    We train ResNet models for a supervised classification task on a small subset of ImageNet. And visualize (a) effective rank across different depths at convergence and (b) training dynamics of effective rank over time for various $\alpha_{\rm min}$.
    In (a) we see that at convergence, our method consistently decreases the feature rank with various depth and, in (b), this pattern is also shown for standard ResNet model at every stage of training. 
   }
    \label{fig:supervised}
\end{figure}

In this experiment, we modify the standard ResNet-18 model to experiment with different depth models.
By default, the ResNet-18 has a total of 8 residual blocks that are equally distributed into 4 layers. 
To increase model depth, we repeat residual blocks in the 3rd layer to obtain models varying between 8 and 16 total layers. At convergence, we observe that the models of different depths achieve a similar test accuracy. 
However, despite similar accuracies, in Figure~\ref{fig:effective_rank_supervised_depth}, which visualizes the effective rank over depth for different values of $\alpha_{\rm min}$, we see that the effective rank decreases over depth. 
Furthermore, smaller values of $\alpha_{min}$ consistently lead to features with lower effective rank.

Next, in Figure~\ref{fig:effective_rank_supervised_time}, we try to verify our conjecture by visualizing the evolution of effective rank during training when choosing different $\alpha_{\rm min}$ in our method.
For this experiment, we choose to train the standard ResNet-18 using our decayed identity shortcuts. In this setup, we observe that the optimal choice of $\alpha_{\rm min}$ slightly improves the test accuracy of the classification network: 94.4\% with $\alpha_{\rm{min}} = 0.7$ \emph{vs.} 93.6\% with $\alpha_{\rm{min}} = 1.0$.
We observe that the effective rank of the final features decreases with decreasing $\alpha_{\rm min}$. This supports our 
hypothesis that (1) decayed identity shortcuts
substantially decrease the rank of bottleneck features and (2) decreasing feature rank may help improve learned features.

\begin{figure*}[t]
    \centering
    \includegraphics[width=\textwidth]{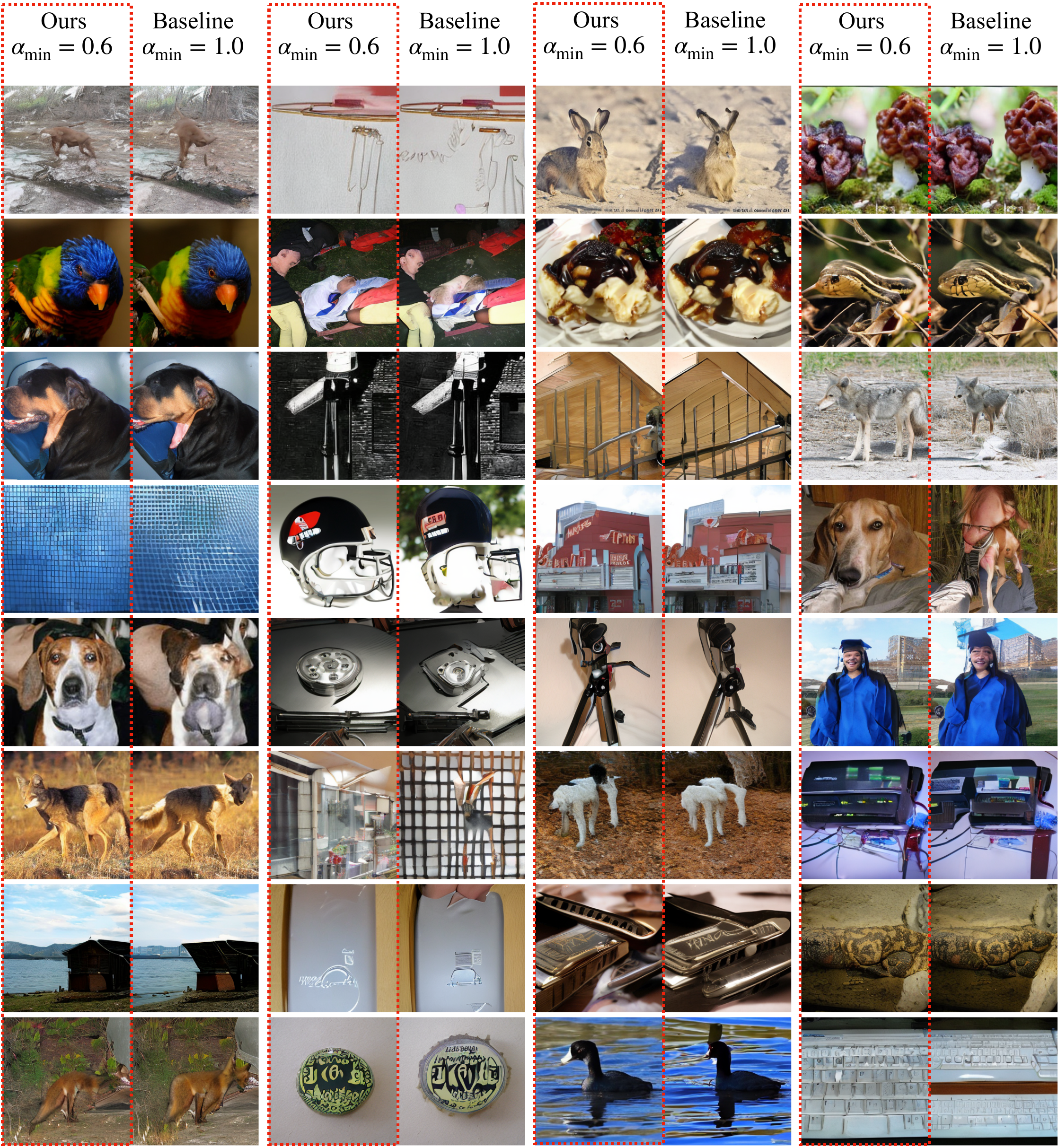}
    \caption{\textbf{Qualitative comparison of images generated by diffusion models.}
    Our method, decayed identity shortcuts with $\alpha_{\rm min} = 0.6$, shows improved representation learning and produces higher-quality generated images compared to the baseline, which employs full residual connections ($\alpha_{\rm min} = 1.0$).
    }
    \label{fig:diffusion_images}
\end{figure*}

\subsection{Further analysis on low-rank property}
\label{app:theoretical}
Consider a deep linear encoder with residual (skip) connections, \( h_{l+1} = \alpha_l\, h_l + W_l\, h_l = (\alpha_l I + W_l)\,h_l \), where \( \alpha_l \in (0,1] \) is a depth-dependent decay factor. The overall encoder mapping is \( H = (\alpha_{L-1}I + W_{L-1})\cdots(\alpha_0 I + W_0) \), a product of perturbed identity transformations. In the standard \( \alpha_l=1 \) case (ResNet), unweighted identity shortcuts cause the output to remain largely a copy of the input, reducing the need to learn complex transformations and yielding high-rank feature representations that preserve fine-grained input details. By contrast, decaying \( \alpha_l<1 \) gradually suppresses this direct copy effect. Expanding the recursion shows that the raw input contribution to \( h_L \) is scaled by the product \( \prod_{i=0}^{L-1}\alpha_i \), whereas contributions involving deeper transformations \( W_i \) omit some of these factors. Thus, for a monotonically decreasing schedule \( \{\alpha_l\} \), the direct “identity” component \( \prod_{i}\alpha_i\, x \) vanishes exponentially with depth. Gradient-based training of deep linear networks is known to converge to minimal-norm solutions, which correspond to mappings with reduced effective rank. Decayed identity shortcuts recover this low-rank inductive bias by progressively weakening the high-rank “copy” component of residual connections, while still preserving the optimization benefits of skip connections for stable training.

 Moreover, the denoising (or masked) autoencoder objective accentuates the effect: since the input has noised (masked) entries, the encoder–decoder must rely on the most salient shared variations in the data (analogous to principal components) to infer missing content, rather than trivially forwarding local details. Thus, the encoder’s output covariance concentrates in a subspace spanned by a few significant singular vectors, and the normalized singular-value entropy is correspondingly low. In other words, the representation has \textit{low effective rank}, which aligns with empirical observations that networks with decayed shortcuts learn features of lower rank that are associated with improved downstream performance.

\end{document}